\documentclass[runningheads]{llncs}
\usepackage[T1]{fontenc}
\usepackage{graphicx}
\usepackage{lineno,hyperref}

\usepackage{graphicx}%
\usepackage{multirow}%
\usepackage{amsmath,amssymb,amsfonts}%
\usepackage{mathrsfs}%
\usepackage[title]{appendix}%
\usepackage{xcolor}%
\usepackage{textcomp}%
\usepackage{manyfoot}%
\usepackage{booktabs}%
\usepackage{algorithm}%
\usepackage{algorithmicx}%
\usepackage{algpseudocode}%
\usepackage{listings}%
\usepackage[figuresright]{rotating}

\usepackage{threeparttable}
\usepackage{makecell}
\usepackage{booktabs}
\usepackage{arydshln}
\usepackage{caption}

\begin{document}
\title{Lighten CARAFE: \\Dynamic Lightweight Upsampling with Guided Reassemble Kernels}
\titlerunning{DLU: Dynamic Lightweight Upsampling}
%
\author{Ruigang Fu\inst{1} \and
Qingyong Hu\inst{2} \and
Xiaohu Dong\inst{1} \and
Yinghui Gao\inst{2} \and
Biao Li\inst{1} \and
Ping Zhong\inst{1}}
\authorrunning{R.G. Fu et al.}
%
\institute{National University of Defense Technology, Changsha 410073, China
\and
Chinese Academy of Military Sciences, Beijing 100000, China \\
\email{\{furuigang08,huqingyong15,dongxiaohu16,yhgao,libiao,zhongping\}@nudt.edu.cn}
\\
}
\maketitle              
\begin{abstract}
As a fundamental operation in modern machine vision models, feature upsampling has been widely used and investigated in the literatures. An ideal upsampling operation should be lightweight, with low computational complexity. That is, it can not only improve the overall performance but also not affect the model complexity. Content-aware Reassembly of Features (CARAFE) is a well-designed learnable operation to achieve feature upsampling. Albeit encouraging performance achieved, this method requires generating large-scale kernels, which brings a mass of extra redundant parameters, and inherently has limited scalability. To this end, we propose a lightweight upsampling operation, termed Dynamic Lightweight Upsampling (DLU) in this paper. In particular, it first constructs a small-scale source kernel space, and then samples the large-scale kernels from the kernel space by introducing learnable guidance offsets, hence avoiding introducing a large collection of trainable parameters in upsampling. Experiments on several mainstream vision tasks show that our DLU achieves comparable and even better performance to the original CARAFE, but with much lower complexity, \textit{e.g.,} DLU requires 91\% fewer parameters and at least 63\% fewer FLOPs (Floating Point Operations) than CARAFE in the case of 16$\times$ upsampling, but outperforms the CARAFE by 0.3\% mAP in object detection. Code is available at \url{https://github.com/Fu0511/Dynamic-Lightweight-Upsampling}.

\end{abstract}

\section{Introduction}
Effective scene parsing and object recognition are key abilities required by most intelligent vision systems to perceive and interact with the real world \cite{vijayakumar2024yolo,zou2024segment}. Recently, deep learning-based approaches have attracted increasing attention of the computer vision community \cite{wang2021tiny,zhu2022rfnet} for their strong capability in representation learning. Despite remarkable research progress has been achieved, it remains challenging to deploy large neural models in practical applications, especially under resource-constrained conditions.

To this end, it is desirable to further explore the lightweight design of neural architecture and operation. In particular, feature upsampling, as one of the fundamental operations in modern deep learning architectures \cite{vijayakumar2024yolo,wang2024yolov9}, has been under-explored compared with other basic operations such as convolution and pooling. Therefore, we dive deep into this area and focus on the lightweight upsampling design in this paper.

Traditional interpolation-based upsampling approaches such as nearest neighbor and bilinear interpolation, have been extensively adopted in classical models for their simplicity and the nature of easy-to-implement  \cite{lin2017feature}. However, interpolation operations fundamentally only consider the spatial adjacency relationship, while failing to capture the rich semantic information encoded in the feature map of CNNs \cite{wang2019carafe}. 
This limitation has motivated researchers to explore learnable upsampling techniques that can introduce additional information through the upsampling operation in a data-driven manner \cite{wang2020deep}. The most commonly used is deconvolution \cite{noh2015learning}, which can be viewed as an inverse operation of convolution. Albeit intuitive, the deconvolution operation has not considered the local variations explicitly in the images, since it applies the same kernel across different locations.

\begin{figure}[t]
\centering
\includegraphics[width=8cm]{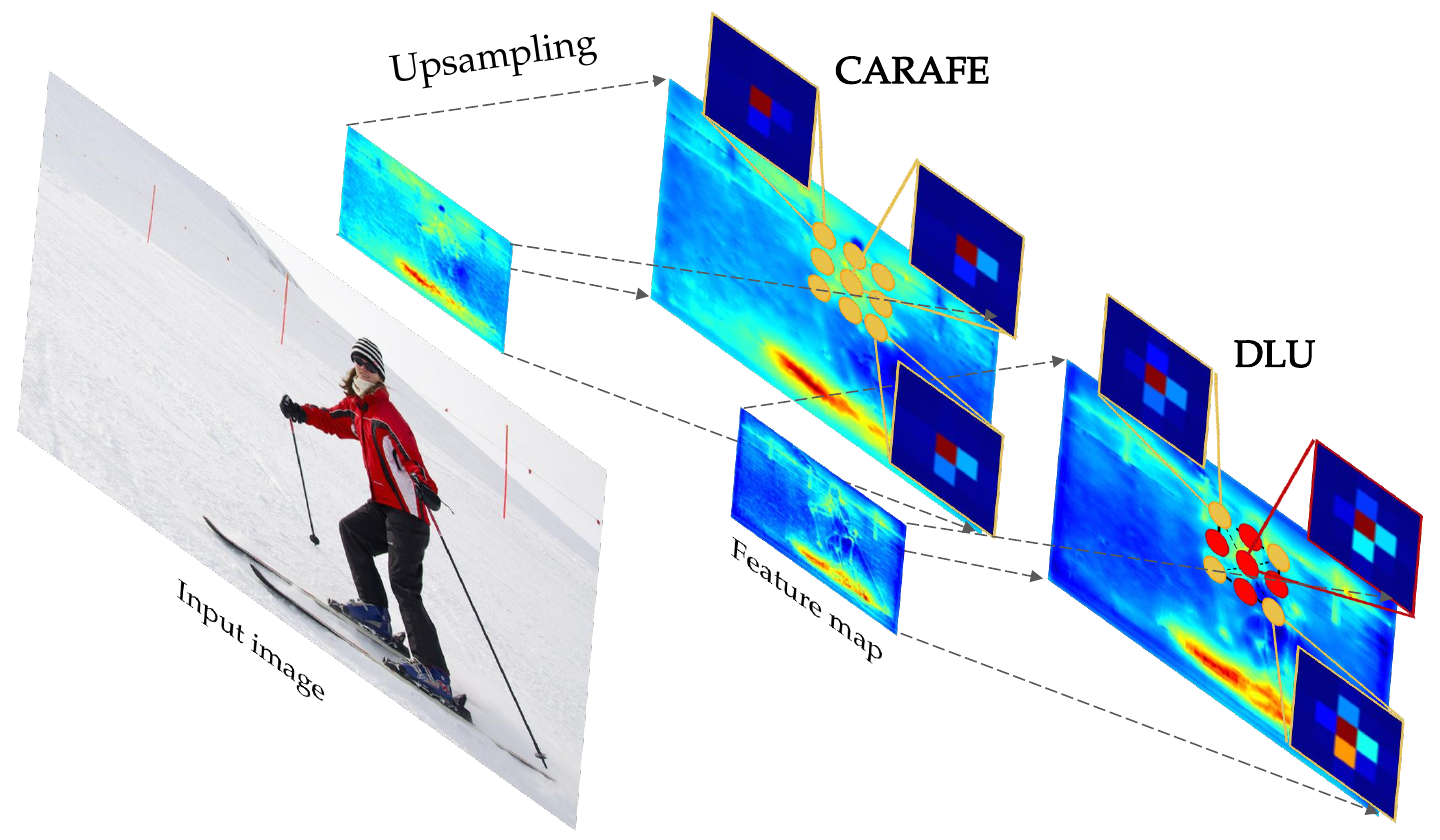}
\caption{Intuitive comparison of CARAFE \cite{wang2019carafe} and DLU. Unlike CARAFE needs to generate independent kernels (yellow ones) for all the positions of the upsampling output, our DLU generates a portion (yellow ones), and samples others (red ones) with learnable offsets through interpolation.}
\label{fig:fig1}
\end{figure}

Compared with the existing interpolation-based and most learning-based upsampling such as deconvolution \cite{noh2015learning} and pixel shuffle \cite{shi2016real}, a learnable upsampling operation named CARAFE (Content-aware Reassembly of Features) \cite{wang2019carafe,wang2021carafe++} achieves promising performance. However, this operation needs to generate a corresponding reassembly kernel for all the positions of the upsampling output, leading to the training parameters being quadratically related to the product of the reassembly kernel size and the upsampling ratio (detail calculation can be found in Table \ref{table1} in section \ref{cwc}). This inherently limits the scalability of the proposed method, especially under large upsampling ratios.

To this end, we proposed a novel upsampling approach termed Dynamic Lightweight Upsampling (DLU) in this paper. As shown in Fig. \ref{fig:fig1}, different from CARAFE generates an independent kernel for each position of the upsampling output, the proposed DLU first constructs a ``source kernel space'' with a portion of kernels, and then samples the kernels with learnable offsets from the kernel space by interpolation. Extensive experiments on several mainstream vision tasks justify the superior performance of the proposed DLU.

The key contributions of our paper are highlighted as follows:

\begin{itemize}
\setlength{\parsep}{0pt} 
\setlength{\topsep}{0pt} 
\setlength{\itemsep}{0pt}
\setlength{\parsep}{0pt}
\setlength{\parskip}{0pt}

\item We propose a novel learnable upsampling operation called DLU by generating large-scale expected kernels from sampling in a ``source kernel space''. It is lightweight, with low computational complexity, and can be seamlessly plugged into various modern neural architectures.

\item The proposed DLU operation is notably lightweight than the original CARAFE operation, but achieve comparable or even better task performance. For example, DLU needs 90\% fewer parameters and at least 63\% fewer FLOPs than the original CARAFE in the case of 16$\times$ upsampling, but outperforms the CARAFE by 0.3\% mAP in object detection.

\item Extensive experiments on three fundemental vision tasks with four representative neural networks verify the availability and scalability of our DLU. For example, by integrating our DLU, the detection performance of FPN \cite{lin2017feature} and Libra RCNN  \cite{pang2019libra} can be improved by 1.2\% and 0.7\% mAP respectively, compared to their original versions.

\end{itemize}
\section{Related Work} \label{Sec:RW}

{Feature upsampling is widely used in vision tasks such as object detection, semantic segmentation and instance segmentation, to increase the resolution of the feature map. In general, it can be divided into interpolation-based and learning-based upsampling. }

\textbf{Interpolation-based Upsampling.} 
{Representative methods include nearest-neighbor interpolation, bilinear interpolation, and bicubic interpolation. To perform interpolation at a specified position, these methods usually first identify a neighborhood of the specific pixel, and then calculate a weighted sum of the specific neighborhood. In general, these methods are fast and easy-to-implement \cite{lin2017feature}. However, the main drawback of the interpolation-based methods is that the output of upsampling is totally derived from the input. That is, no additional information was introduced during the upsampling operation.
}

{ \textbf{Learning-based Upsampling.} Different from interpolation-based upsampling, learning-based upsampling methods aim to introduce additional information during the upsampling operation, hence attracting increasing attention in recent years. Deconvolution operation \cite{noh2015learning} is proposed to perform zero-padding of the input feature map first, and then increase the feature resolution by executing standard convolutions. However, the effective receptive field of deconvolution is rather small. Pixel shuffle \cite{shi2016real} achieves upsampling by first increasing the feature channels through convolution, and then reshaping to a specific resolution. This method can achieve a larger receptive field, but with the price of massive extra parameters. To improve the upsampling efficiency, Mazzini et al. \cite{mazzini2018guided} proposed guided upsampling, which enlarges the input using bilinear interpolation, but sampling pixels with learnable offsets. However, this method also suffers from small effective receptive fields. Building upon guided upsampling, DySample \cite{liu2023learning} delves deeper into the generation of the offsets, exploring superior methods to achieve improved performance.
} 

{A handful recent works have started to explore the upsampling with learnable kernels to dynamically enlarge the receptive field \cite{lu2022sapa,wang2019carafe,lu2022fade}. Among them, CARAFE \cite{wang2019carafe} is an effective learnable upsampling operation. Given the upsampling ratio as $\sigma$, CARAFE would generate $\sigma^2$ kernels for each pixel of the input, then one pixel would output $\sigma^2$ pixels through convoluting the $\sigma^2$ different generated kernels. Later, CARAFE++ \cite{wang2021carafe++} is further proposed to endow the learnable downsampling capacity, following the similar concepts of CARAFE. FADE \cite{lu2022fade} and SAPA \cite{lu2022sapa} refined the framework of CARAFE to a double-input framework. Their input requires not only the low-resolution feature to be upsampled but also a high-resolution guiding feature as an additional input. Consequently, their application scenarios is limited (high-resolution features must be available). Our goal is to contribute a lightweight, scalable and effective upsampling operation. Since CARAFE-based techniques require a large collection of parameters in generating kernels, we lighten the kernel generation module of CARAFE and propose a dynamic lightweight upsampling operator.}

\section{Method}
\subsection{Revisit CARAFE}

{Given a feature map $\mathcal{F}$ of size $H \times W \times C$, and the upsampling ratio as $\sigma$, the expected size of outputted feature map after upsampling is $\sigma H \times \sigma W \times C$. Specifically, CARAFE achieves feature upsampling through the following steps: 1) Predicting a tensor of size $H \times W \times k_{up}^2\sigma^2$ by convoluting the input feature map, where  $k_{up} \times k_{up}$ is the receptive field (kernel size) for upsampling. This tensor is further reshaped to $\sigma H \times \sigma W \times k_{up}^2$ according to pixel shuffle \cite{shi2016real}, and each spatial location in the tensor forms a $k_{up} \times k_{up}$ kernel, resulting in $\sigma H \times \sigma W$ kernels in total (known as reassemble kernels). 2) All these kernels are normalized by softmax function, respectively. 3) For each location of the expected output $(i,j,c)$ (position $(i,j)$ in the $c$-th feature map), taking the predicted kernel at position $(i,j)$ from the above kernel tensor, and finding the corresponding neighbourhood of $k_{up} \times k_{up}$ around the input pixel $(\lfloor i/\sigma \rfloor, \lfloor j/\sigma \rfloor, c)$. The final response at $(i,j,c)$ of the upsampled feature maps is the inner product of the kernel and the pixel neighborhood.
}

{\textbf{Limitation of CARAFE.} CARAFE uses a convolution layer to predict a tensor of size $H \times W \times k_{up}^2\sigma^2$ in the branch of reassemble kernel generation, where the parameters in this convolution layer are proportional to $k_{up}^2\sigma^2$. This comes with a large collection of trainable parameters, resulting in memory-inefficient, especially when a large receptive field (\textit{i.e.,} large $k_{up}$) or aggressive upsampling ratio (\textit{i.e.,} large $\sigma$) is used. Additionally, too many parameters also lead to a high risk of overfitting.} 

\textbf{Solutions.} 

It is important to note that the kernels predicted by CARAFE are independent of each other. However, it is argued that using independent kernels to convolute an input may lead to an issue of so-called ``checkerboard artifacts'' \footnote{\url{https://distill.pub/2016/deconv-checkerboard/}}, since there is no direct relationship between adjacent pixels in the upsampled output \cite{gao2019pixel}. As a result, the large collection of trainable parameters in CARAFE are likely to be redundant. To address this issue, it may be more feasible and effective to explore the interdependence of adjacent kernels.

Inspired by the idea of deformable kernels \cite{gao2019deformable}, that is, kernels can be augmented by learning free-form offsets on kernel coordinates in the original kernel space, we further proposed a lightweight upsampling operation by avoiding generating a large collection of independent kernels. Specifically, we first construct a small-scale source kernel space, and then sample large-scale expected kernels from the source kernel space by introducing learnable guidance offset. It is noted that the trainable parameters can be greatly reduced in this way.

\begin{figure}[t]
\centering
\centerline{\includegraphics[width=12cm]{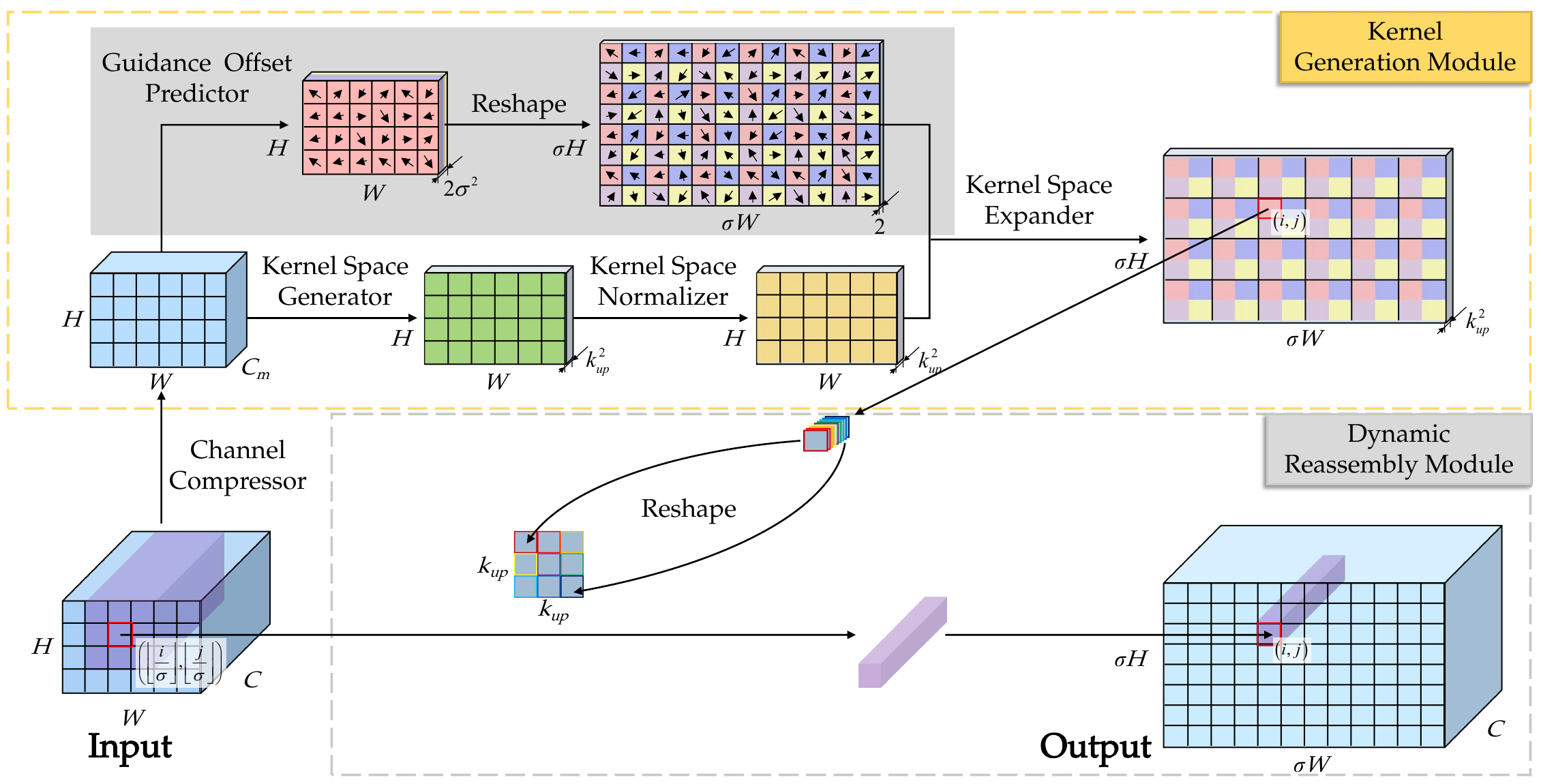}}
\caption{The overall framework of our DLU.}
\label{fig:fig2}
\end{figure}

\subsection{The Proposed DLU}

{As shown in Fig. \ref{fig:fig2}, the proposed DLU mainly consists of two modules: kernel generation and dynamic reassembly module. Given a scaling ratio $\sigma$ for feature upsampling, the kernel generation module is used to predict $\sigma H \times \sigma W$ kernels, where $H$ and $W$ determine the shape of the input feature map. The dynamic reassembly module is responsible for {generating the upsampling output with shape of $\sigma H \times \sigma W$}, each location of the input is convoluted with location-specific generated kernels \cite{jia2016dynamic}. That is, the convolution is dynamic and content-aware.}

\subsubsection{Kernel Generation Module}
{Different from the original CARAFE \cite{wang2019carafe}, the kernel generation module in our framework is composed of: channel compressor, kernel space generator, kernel space normalizer, guidance offset predictor, and kernel space expander.}

\textit{\textbf{Channel Compressor.}} 
{Similar to CARAFE, this step aims to reduce the number of input feature channels, hence further saving the overall computational cost. Specifically, we compress the input feature map from $C$ channels to $C_m$ channels through a 1$\times$1 convolution layer. }

\textit{\textbf{Kernel Space Generator.} }
Next, a $k_{encoder} \times k_{encoder}$ convolution layer is used to encode the {compressed feature maps} to generate a tensor of size $H \times W \times k_{up}^2$. Each location in the tensor can be reshaped as a $k_{up} \times k_{up}$ kernel, where $k_{up}$ determined the size of the receptive field for upsampling. Up to here, the tensor with $H \times W$ kernels forms a ``source kernel space''.

\textit{\textbf{Kernel Space Normalizer.} }
{Here, the kernel space is normalized by performing softmax operation spatially. As a result, the elements for each kernel in the space are forced to be positive, with a summation of 1.} This step is motivated by the classical interpolation-based upsampling methods. Specifically, our DLU can be fundamentally regarded as a linear weighted method, \textit{i.e.,} each position of the upsampled feature is a weighted sum of the input feature and the generated kernels represent the weights. Different from traditional deconvolution operation, softmax normalization of the kernels is used to make the upsampled feature have the same order of magnitudes as the input feature, {which provides clearer physical interpretability.}

\textit{\textbf{Guidance Offset Predictor.} }
Here, the proposed guidance offset predictor is used to predict the sampling coordinates for enlarging the kernel space. Specifically, a $k_{encoder} \times k_{encoder}$ convolutional layer is used to convolute the compressed feature map to generate a tensor of size $H \times W \times 2\sigma^2$. This tensor can be further reshaped to $\sigma H \times \sigma W$ 2-D offsets (in $x$ and $y$ dimensions respectively), as shown in Fig. \ref{fig:fig2}.

\begin{figure}[t]
\centering
\centerline{\includegraphics[width=11cm]{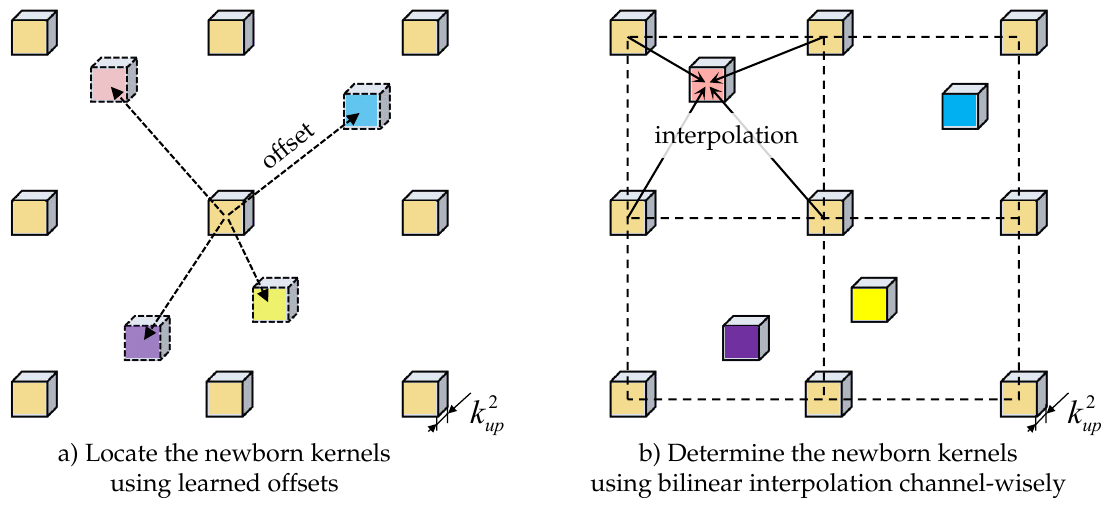}}
   \caption{{Illustration of the expansion of source kernel space. In this figure, each kernel in the source space (orange) generates 4 newborn kernels (pink, blue, purple, and yellow) in the case of 2$\times$ upsampling, and the newborn kernels are sampled from the source kernel space using bilinear interpolation channel-wisely.}}
\label{fig:fig3}
\end{figure}

\textit{\textbf{Kernel Space Expander.} }
For each location in the normalized source kernel space, we assign $\sigma^2$ 2-D offsets and sample $\sigma^2$ newborn kernels by applying bilinear interpolation channel-wisely. Specifically, Fig. \ref{fig:fig3} illustrates the expansion of the source kernel space.

Overall, the kernel generation module results in $\sigma H \times \sigma W$ final kernels, as shown in Fig. \ref{fig:fig2}. Note that it can be proved that arbitrary kernel in the expanded kernel space is also softmax normalized after the normalization of the source kernel space (The detailed proof can be found in our appendix).

\subsubsection{Dynamic Reassembly Module}
Following the original CARAFE \cite{wang2019carafe}, a dynamic reassembly module is used to output the upsampled feature map which is expected to have the size of $\sigma H \times \sigma W \times C$. To get the response at position $(i,j)$ in the $c$-th feature map of the expected output, we first take the predicted kernel at position $(i,j)$ from the expected expanded kernel space, and then dig out a neighborhood of $k_{up} \times k_{up}$ around the input pixel at position $(\lfloor i/\sigma \rfloor, \lfloor j/\sigma \rfloor)$ in the $c$-th feature map of the input, then, the specific response in the output is equal to the inner product of the kernel and the pixel neighborhood.

\subsection{Comparison with CARAFE} \label{cwc}

{As described above, the key difference between CARAFE and our DLU is the way to generate reassemble kernels, which further leads to different number of trainable parameters and computational costs.

Based on \cite{molchanov2016pruning} and \cite{wang2019carafe}, given the task of upsampling a feature map with input channel $C_{in}$ by a ratio of $\sigma$, the details of the required parameters and computation in the kernel generation module of CARAFE [18] and our DLU can be found in Table \ref{table1}. Table \ref{table1} shows that both the parameters and computational complexity of CARAFE are primarily proportional to $\sigma^2k_{up}^2$, while in our DLU, the parameters and computational complexity are primarily proportional to $k_{up}^2+2\sigma^2$. This indicates that our DLU model generally has fewer parameters and lower computational complexity than CARAFE when $k_{up}\geq3$ and $\sigma\geq2$ (general cases). We also noticed that this advantage of our approach is more distinguished if dense upsampling with aggressive upsampling ratios or large receptive fields are used.

Note that, our DLU operator introduces bilinear interpolation in its kernel space expander, which results in a term of $9\sigma^2k_{up}^2$ being included in the total computation. However, this term is typically negligible compared to the computation of CARAFE. This is because the kernel generator of CARAFE requires $2(C_mk_{encoder}^2+1)\sigma^2k_{up}^2$ FLOPs, and $C_mk_{encoder}^2$ is typically a large value.

\begin{table}[!htbp]
\caption{Details of the required parameter and computation in the kernel generation module of CARAFE \cite{wang2019carafe} and our DLU.}
\centering
{\begin{threeparttable}
{
\scalebox{0.9}{
\begin{tabular}{cc}
\toprule
{\# Trainable Parameters} & {Kernel Generation Module}\\
\toprule
CARAFE & $\underbrace{(C_{in}+1)C_m}_{Channel\ Compressor}$+ $\underbrace{(C_mk^2_{encoder} + 1)\boldsymbol{\sigma^2k^2_{up}}}_{Kernel\ Generator}$\\
 \cdashline{1-2}
DLU & \makecell[c]{$\underbrace{(C_{in}+1)C_m}_{Channel\ Compressor}$+ $\underbrace{(C_mk_{encoder}^2+1)\boldsymbol{k_{up}^2}}_{Kernel\ Space\ Generator}$+$\underbrace{(C_mk_{encoder}^2+1)\boldsymbol{2\sigma^2}}_{Guidance\ Offset\ Predictor}$}\\

 \hline 
 \midrule
 {\# Per-pixel FLOPs}&{Kernel Generation Module}\\
\midrule
 CARAFE &$\underbrace{2(C_{in}+1)C_m}_{Channel\ Compressor}$+$\underbrace{2(C_mk^2_{encoder} + 1)\boldsymbol{\sigma^2k^2_{up}}}_{Kernel\ Generator}$+ $\underbrace{\boldsymbol{\sigma^2} \times (k_{up}^2\rm {-D\ sm})\tnote{1}}_{Kernel\ Normalization}$\\ 
 
 \cdashline{1-2}
 
 DLU &\makecell[c]{$\underbrace{2(C_{in}+1)C_m}_{Channel\ Compressor}$+$\underbrace{2(C_mk_{encoder}^2+1)\boldsymbol{k_{up}^2}}_{Kernel\ Space\ Generator}$+ $\underbrace{\boldsymbol{1} \times (k_{up}^2\rm {-D\ sm})}_{Kernel\ Normalization}$\\+$\underbrace{2(C_mk_{encoder}^2+1)\boldsymbol{2\sigma^2}}_{Guidance\ Offset\ Predictor}$+$\underbrace{\boldsymbol{9\sigma^2k^2_{up}}}_{Kernel\ Space\ Expander}$}\\

\bottomrule
\end{tabular}
}}
\scriptsize\begin{tablenotes}
\item [1] ($k_{up}^2$-D sm) represents the computation of softmax normalization for a $k_{up}^2$-D vector.
\end{tablenotes}
\end{threeparttable}}
\label{table1}
\end{table}

\section{Experiments and Results} \label{Sec:Ex}

In this section, we mainly evaluate the performance of different upsampling methods on multiple mainstream vision tasks. Specifically, these upsampling methods are integrated into several typical deep-learning models to replace the original upsampling operators. 
\subsection{Experimental Setup}
\begin{itemize}
\setlength{\parsep}{0pt} 
\setlength{\topsep}{0pt} 
\setlength{\itemsep}{0pt}
\setlength{\parsep}{0pt}
\setlength{\parskip}{0pt}
\item \textbf{Object Detection.} For object detection, we integrate different upsampling operations into the FPN \cite{lin2017feature} framework, and then compare the performance on the MS COCO dataset \cite{lin2014microsoft}. Results are evaluated based on the standard COCO metric, \textit{i.e.} mAP of IoUs from 0.5 to 0.95. Besides, we also have experiments on Libra CNN \cite{pang2019libra}, evaluating the performance of the original CARAFE and our proposed methods with different upsampling ratios. 

\item \textbf{Semantic Segmentation.} 
We take Semantic FPN \cite{kirillov2019panoptic} as the baseline framework and then compare the semantic segmentation performance of different upsampling operations on the ADE20K dataset \cite{zhou2017scene}. Results are measured with mean IoU (mIoU) which indicates the average IoU between predictions and ground truth masks.

\item \textbf{Instance Segmentation.} 
For instance segmentation, we take the Mask RCNN \cite{He2017Mask} as the baseline framework and then compare the segmentation performance of different upsampling operations on the MS COCO dataset. Due to the page limits, please refer to our appendix for the experimental results on instance segmentation.

\item \textbf{Implementation Details.} 
All the deep models are trained on the training subset, and the final performance is evaluated on the validation set. Experiments of object detection and instance segmentation are implemented based on the mmdetection toolbox \cite{mmdetection} and follow the 1$\times$ training schedule settings on four NVIDIA Quadro P5000 GPUs. Specifically, for training our models, we employed stochastic gradient descent for 12 epochs with a batch size of 2 examples per GPU, weight decay of 0.0001, momentum of 0.9, and a learning rate of 0.0025. The initial learning rate decayed by a factor of 10 at the 8$^\mathrm{th}$ and 11$^\mathrm{th}$ epochs. Similarly, for our experiments on semantic segmentation, we used the mmsegmentation toolbox \cite{mmseg2020} and followed the 40k training schedule settings on four NVIDIA Quadro P5000 GPUs. Our models were trained using stochastic gradient descent for 40k iterations with a batch size of 4 examples per GPU, weight decay of 0.0005, momentum of 0.9, and a learning rate of 0.01. We utilized the poly learning rate policy with a power factor of 0.9 to decay the initial learning rate.

\end{itemize}

Unless noted, our DLU shares the same hyper-parameters with CARAFE in all the experiments. Specifically, $C_m$ = 64 for the channel compressor, $k_{encoder}$ = 3, $k_{up}$ = 5 for the kernel space generator. The ablation studies of hyperparameters impact can be found in the appendix. We initialized the weights of the kernel space generator from a zero-mean Gaussian distribution with standard deviation of 0.001, {since a large standard deviation may lead to small gradients in backpropagation for the following kernel space normalizer (softmax function).} Then, the weights of the guidance offset predictor are initialized by zero-initializing, while other layers follow adaptive Xavier initialization.

\begin{figure}[t]
\centering
\centerline{\includegraphics[width=8cm]{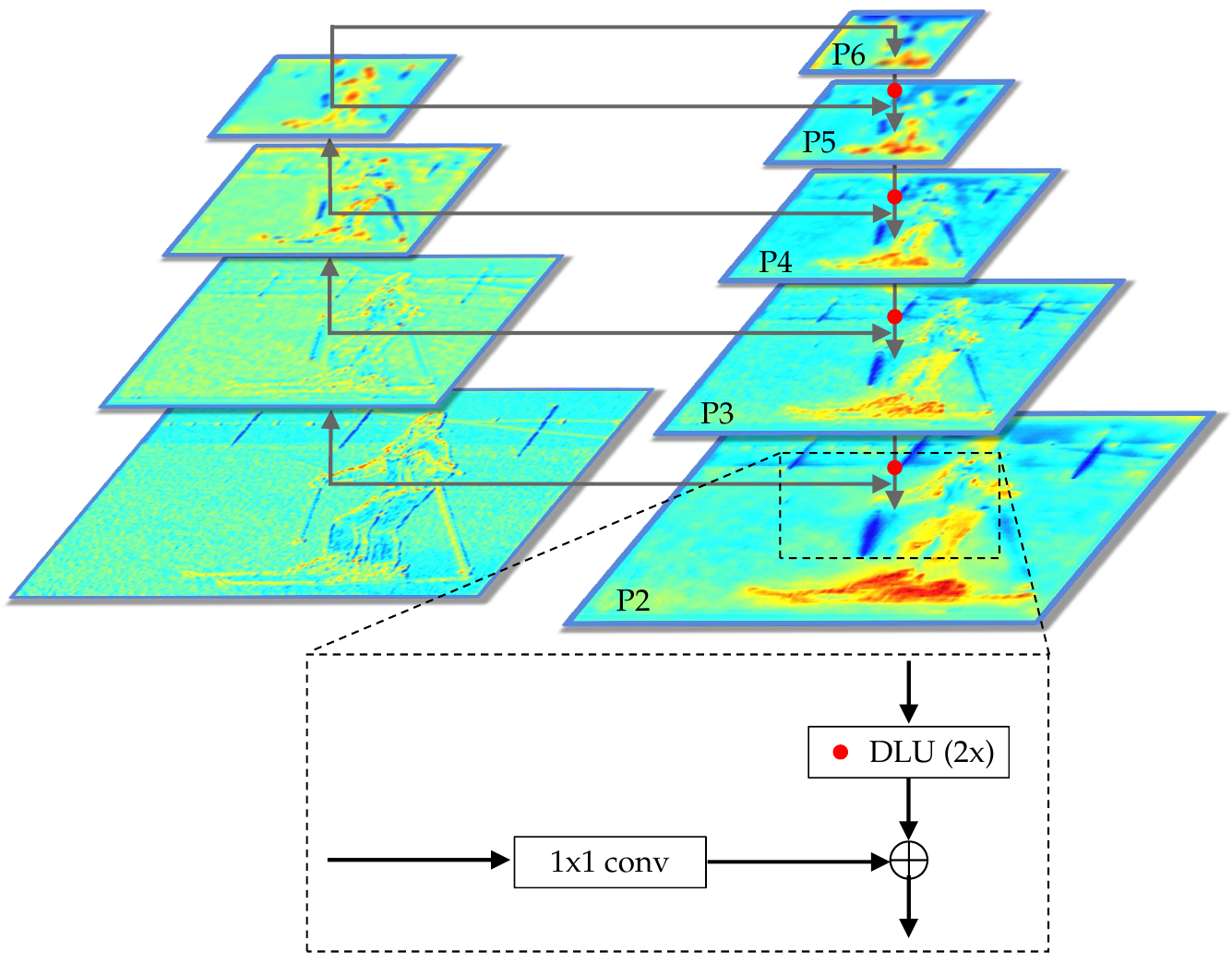}}
\caption{FPN architecture with the upsampling operator as DLU.}
\label{fig:fig5}
\end{figure}

\subsection{Object Detection}
\label{sec::Object}
\subsubsection{FPN}
{
We take the FPN \cite{lin2017feature} as the backbone. Specifically, FPN constructs a feature pyramid to exploit multi-scale features with a top-down pathway and several lateral connections. As shown in Fig. \ref{fig:fig5}, the lateral connections first are used to align the feature channels by applying a $1 \times 1$ convolution. Then, features of different scales are fused in the top-down pathway by upsampling the low-resolution features by 2$\times$.}

In our experiments, we adopted a feature pyramid of 5 feature levels: \{$P2$, $P3$, $P4$, $P5$, $P6$\} with strides \{4, 8, 16, 32, 64\}, and 4 upsampling operators with a scaling ratio of 2 are used totally, as shown in Fig. \ref{fig:fig5}. To compare the impact of different upsampling operations on object detection performance, we first replace the original nearest-neighbor interpolation with different upsampling operators in FPN, and then evaluate the detection performance. Table \ref{table2} shows the quantitative detection results achieved by FPN with different upsampling operators on the MS COCO 2017 validation set. We analyse the results from the following aspects:

\begin{table}[!t]
\caption{{Detection results on MS COCO 2017 validation set with FPN \cite{lin2017feature}. Various upsampling operations are used. Please note that ``Params'' represents the introduced additional parameters per upsampling operation, ``FLOPs'' represents the FLOPs of each upsampling operation corresponding to each pixel of the input feature map. Above the dotted line are interpolation-based methods, while below the dotted line are learning-based methods. Best performance is in boldface, and second best is underlined.}} \label{table2}
\centering
\begin{threeparttable}

\scalebox{0.88}{
\begin{tabular}{rccccccccc}
\Xhline{2.0\arrayrulewidth}
Method & mAP & mAP$_{50}$ & mAP$_{75}$ & mAP$_{S}$ & mAP$_{M}$ & mAP$_{L}$ & Params & FLOPs & FPS \\
\Xhline{2.0\arrayrulewidth}
Nearest & 37.4 & 58.1 & 40.6 & 20.9 & 41.2 & 48.9 & 0 & 0 &{20.4}\\
Bilinear & 37.5 & 58.5 & 40.5 & 21.3 & 41.1 & 48.8 & 0 & 9K & 20.3\\
 \cdashline{1-10}
\multicolumn{10}{c}{double-input flow} \\
FADE\cite{lu2022fade} & 38.5 &59.6 &{\bf 41.8} &{\bf 23.1} &{\underline {42.2}} &49.3&49K&-&-\\
SAPA-B\cite{lu2022sapa} & 37.8 &59.2 &40.6 &22.4 &41.4 &49.1&{\bf 25K}&-&-\\
\multicolumn{10}{c}{single-input flow} \\
Deconv\cite{noh2015learning} & 37.2 & 57.8 & 40.2 & 20.5 & 41.0 & 48.0 & 262K & 1.2M & {\underline {18.8}} \\
Pixel shuffle\cite{shi2016real}  & 37.6 & 58.5 & 40.6 & 21.1 & 41.4 & 48.3 & 2.4M & 4.7M & 18.2  \\
CARAFE\cite{wang2019carafe} & {\bf 38.6} & {\bf 59.9} & 41.6 & {\underline {23.0}} & {\bf 42.3} & {\underline {49.5}} & 74K & {\underline {199K+4$\times$(25-D sm)\tnote{1}}} & {\bf 19.4} \\
DLU(Ours) & {\bf 38.6} & {\bf 59.9} & {\bf 41.8} & {22.6} & {\underline {42.2}} & {\bf 50.1} & {\underline {35K}} & {\bf 123K+1$\times$(25-D sm)} & {\underline {18.8}}  \\

\Xhline{2.0\arrayrulewidth}
\end{tabular}}
\scriptsize\begin{tablenotes}
\item [1] (25-D sm) represents the computation of softmax normalization for a 25-D vector.
\end{tablenotes}
\end{threeparttable}
\end{table}

(1) \textbf{Effectiveness}. By simply replacing the original nearest-neighbor interpolation with the proposed DLU in the framework of baseline FPN, the best detection performance is achieved, with a 1.2\% improvement on mAP. We can also see that the proposed operation is beneficial for the detection performance with various object scales, since the AP$_S$, AP$_M$, AP$_L$ (small, medium, large) are all improved. More quantitative results can be seen in Fig. \ref{fig:fig7}.

(2) \textbf{Scalability (Memory Consumption)}. FADE\cite{lu2022fade} and SAPA\cite{lu2022sapa} are methods with double-input flow, that is, their input not only comprises the low-resolution feature to be upsampled, but also needs a high-resolution guiding feature. While they also successfully lighten CARAFE, their dependency on the availability of the high-resolution guiding features restricts their application scenarios.

For methods with single-input flow, identical detection performance is achieved when the original CARAFE or the proposed DLU is used in the FPN framework. However, the proposed DLU is much more lightweight compared with the original CARAFE. The detailed quantitative results in our case are shown in Table \ref{table2}, 74K additional parameters are required per CARAFE, while our DLU only requires 35K additional parameters, up to half fewer parameters, but with comparable performance. This further indicates that the parameters of the original CARAFE are redundant. The reduction of redundancy in parameters makes the computation complexity of DLU lower than the original CARAFE in terms of FLOPs ($\approx$ 123k vs. 199k in our case).


\begin{figure}[t]
\centering
\centerline{\includegraphics[width=10.5cm]{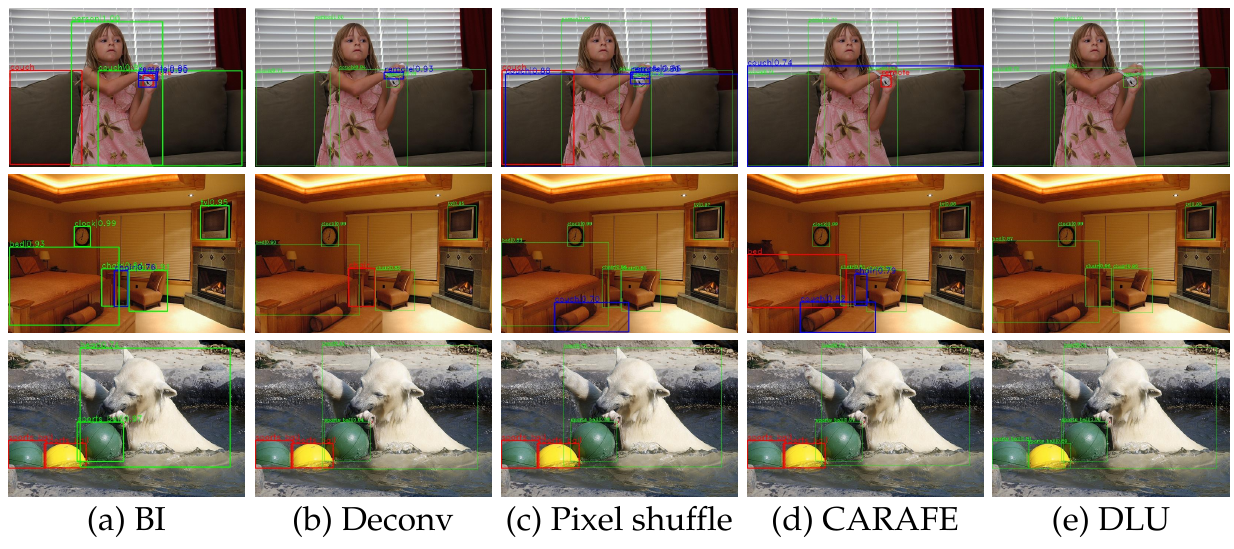}}
\caption{Comparison of detection results of FPN \cite{lin2017feature} with different upsampling methods on the validation set of COCO 2017. The true positives, false
positives, and false negatives are indicated by \textcolor{green}{green}, \textcolor{blue}{blue}, and \textcolor{red}{red} rectangles, respectively. ``BI'' represent for Bilinear Interpolation.}
\label{fig:fig7}
\end{figure}

(3) \textbf{Efficiency}. DLU has lower FLOPs than CARAFE, however, we also found that DLU's actual FPS is not improved compared to CARAFE. We attribute this to the inferior degree of parallelism in the interpolation operation in a GPU environment. Nevertheless, we are confident that with increased parallelism and code optimization in the future work, our method will undergo significant improvements in terms of efficiency.

\subsubsection{Libra RCNN}

Libra RCNN \cite{pang2019libra} is another representative neural architecture in object detection. It explores better utilization of the multi-level features of FPN \cite{lin2017feature} to generate more discriminative pyramid representations. Specifically, Libra RCNN proposed a BFP (Balanced Feature Pyramid) module, which first integrates the multi-level outputs of FPN by using multiple rescaling operations and element-wise summation. Then, it refines the so-called balanced semantic features by several refinement operations (\textit{e.g.,} convolution). Finally, the obtained feature is rescaled back to the pyramid resolutions and then added to the original FPN features. In our experiments, we build the BFP on a FPN with 5 output feature maps and rescale all features to the highest resolution. That is, 4 upsampling operations with ratio of 2$\times$, 4$\times$, 8$\times$ and 16$\times$ are needed in our BFP.

\begin{table}[t]
\caption{Detection results achieved by Libra CNN \cite{pang2019libra} on MS COCO 2017 validation set with different upsampling operations. {Note that the default upsampling operation (untick) used in the BFP module is the nearest neighbor interpolation.}}\label{table3}
\centering
\tabcolsep=0.3cm
\footnotesize\begin{tabular}{ccccccc}
\Xhline{2.0\arrayrulewidth}
{Method} & {$\times$2} & {$\times$4} & {$\times$8} & {$\times$16} & {mAP} & {\textbf{Params}} \\
\Xhline{2.0\arrayrulewidth}
{Baseline} & {} & {} & {} & {} & {37.7} & {0} \\
\cdashline{1-7}
\multirow{5}{*}{CARAFE} 
    & {\checkmark} & {} & {} & {} & {37.9}& {74K} \\
    & { } & {\checkmark} & {} & {} & {37.8} & {247K}\\
    & { } & {} & {\checkmark} & {} & {38.1} & {939K}\\
    &{ } & { } & {} & {\checkmark} & {38.1} & {3.7M} \\
    & {\checkmark} & {\checkmark} & {\checkmark} & {\checkmark} & {38.1}& {4.9M} \\
\cdashline{1-7}
\multirow{5}{*}{DLU} 
    & {\checkmark} & {} & {} & {} & {37.9}& {35K} \\
    & { } & {\checkmark} & {} & {} & {38.0} & {49K}\\
    & { } & {} & {\checkmark} & {} & {38.2} & {104K}\\
    &{ } & { } & {} & {\checkmark} & {38.3} & {326K} \\
    & {\checkmark} & {\checkmark} & {\checkmark} & {\checkmark} & {\bf 38.4}& {515K} \\

\Xhline{2.0\arrayrulewidth}
\end{tabular}
\end{table}

\begin{figure}[ht]
\centering
\centerline{\includegraphics[width=9cm]{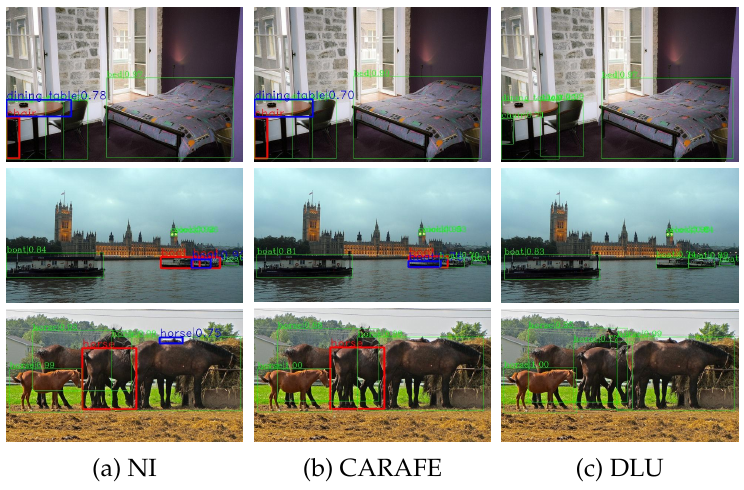}}
\caption{Comparison of detection results of Libra RCNN \cite{pang2019libra} with different upsampling methods on the validation set of COCO 2017. The true positives, false
positives, and false negatives are indicated by \textcolor{green}{green}, \textcolor{blue}{blue}, and \textcolor{red}{red} rectangles, respectively. ``NI'' represent for Nearest-neighbor
Interpolation.}
\label{fig:fig8}
\end{figure}

{To fairly compare the performance of CARAFE and our DLU under different upsampling ratios, we only replace the upsampling layer in BFP, while keeping the rest of FPN unchanged. As shown in Table \ref{table3}, the proposed DLU improve the baseline by 0.7\% mAP, and is consistently better than CARAFE under different settings.} Furthermore, our DLU requires far fewer parameters than CARAFE, which is more visible at large sampling rates (\textit{e.g.,} 326K vs. 3.7M in the case of 16$\times$
upsampling). This further verifies that the parameters in CARAFE are indeed redundant, and excessively redundant parameters also increase the risk of model overfitting (\textit{i.e.,} the model parameters increase but the performance decreases). 

We conducted a qualitative comparison of the detection performance achieved by Libra RCNN by replacing all the upsampling operations in the BFP with different methods. The results are shown in Fig. \ref{fig:fig8}.

\subsection{Semantic segmentation}
To further evaluate the effectiveness of our proposed DLU, we conducted experiments on the task of semantic segmentation. These experiments provide additional insights into the applicability and potential benefits of DLU in other computer vision tasks beyond object detection.

For the task of semantic segmentation, we adopted the representative Semantic FPN model \cite{kirillov2019panoptic} as our baseline network. The architecture of Semantic FPN is based on the FPN model \cite{lin2017feature}. Specifically, in its module of semantic segmentation branch, each FPN level is upsampled using convolutions and 2$\times$ bilinear upsampling progressively until it reaches a 1/4 scale of the input. These outputs are then fused using summation operation and finally decoded to a pixel-wise output using a 4$\times$ bilinear upsampling.

In the above processing, upsampling operations are used for two modules: FUSE (to fuse feature maps with different spatial resolution) and Decoder (to decode the feature map to a pixel-wise output). In our experiments, we replaced the original bilinear interpolation used in these two modules with different upsampling operators and compared their segmentation performance.

\begin{table}[t]
\caption{Segmentation results on ADE20K val with Semantic FPN \cite{kirillov2019panoptic}. Various upsampling methods are used in its FUSE and Decoder modules. The default upsampling operation (untick) is the bilinear interpolation.}\label{table4}
\centering
\tabcolsep=0.3cm
\footnotesize{\begin{tabular}{ccccc}
\Xhline{2.0\arrayrulewidth}
Method & FUSE($\times$2,$\times$4,$\times$8) & {Decoder($\times$4)} & mIoU &Params\\
\Xhline{2.0\arrayrulewidth}
{Baseline} & {} & {} & {35.53} & {0} \\
\cdashline{1-5}
\multirow{2}{*}{CARAFE} 
    & {\checkmark} & {} & {36.32} & {1.2M } \\
    & { } & {\checkmark} & {36.57} & {239K}\\
\cdashline{1-5}
\multirow{2}{*}{DLU} 
    & {\checkmark} & {} & {\bf 37.07} & {165K } \\
    & { } & {\checkmark} & {36.58} & {41K }\\
\Xhline{2.0\arrayrulewidth}
\end{tabular}}
\end{table}

\begin{figure}[t]
\centering
\centerline{\includegraphics[width=10.5cm]{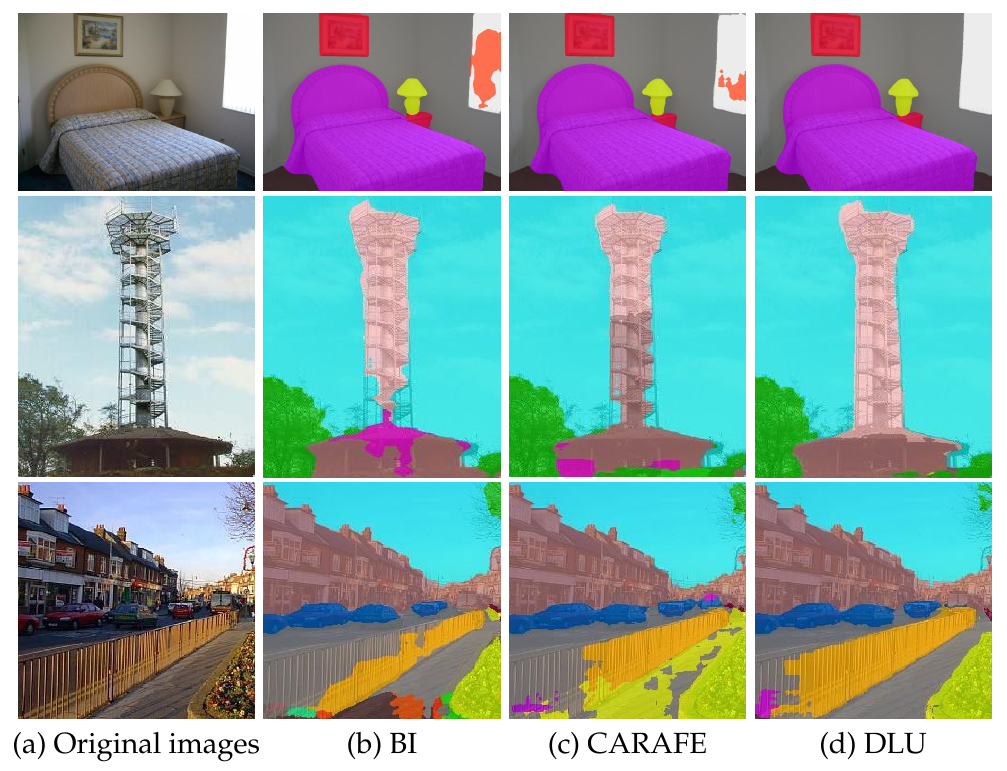}}
\caption{Comparison of segmentation results of Semantic FPN \cite{kirillov2019panoptic} on ADE20K val. Different upsampling methods are used in its FUSE module.}
\label{fig:fig9}
\end{figure}



To compare the performance of CARAFE and our proposed DLU under different upsampling ratios in the context of semantic segmentation, we made a simple modification to the FUSE module. Specifically, instead of enlarging the size of feature maps using convolutions and 2$\times$ upsampling progressively, we used one convolution layer and one upsampling operation with a rather large scaling ratio. For example, for feature maps that are 1/32 scale of the input, to reach 1/4 scale of the input, we first applied a $3\times3$ convolution layer on the feature maps and then resized the feature maps using an 8$\times$ upsampling operation directly.

Our experiments utilized semantic FPN with four feature levels, and we used three upsampling operations with ratios of 2$\times$, 4$\times$, and 8$\times$ in the FUSE module. Table \ref{table4} presents the quantitative semantic segmentation performance achieved by semantic FPN with different upsampling operations.
The results show that, by replacing the baseline bilinear interpolation with DLU in either the FUSE or the Decoder module, the overall performance of semantic segmentation can be significantly improved. For instance, in the FUSE module, DLU outperformed the bilinear interpolation by 1.5\% in terms of mIoU scores and also achieved better task performance than CARAFE, despite having fewer parameters. Figure \ref{fig:fig9} provides a qualitative comparison of segmentation performance achieved by semantic FPN with different sampling operations in its FUSE modules.

\subsection{Discussion}

Albeit much lower complexity compared with CARAFE, we also noticed that the overall framework of our DLU is still similar to the CARAFE. However, this does not mean that our DLU has only limited significance. Instead, the practical significance of our DLU lies in:

\begin{itemize}
\setlength{\parsep}{0pt} 
\setlength{\topsep}{0pt} 
\setlength{\itemsep}{0pt}
\setlength{\parsep}{0pt}
\setlength{\parskip}{0pt}

\item \textbf{Versatility}. As a fundamental operation, upsampling is widely used in most deep neural networks. By replacing the original nearest-neighbor interpolation or bilinear interpolation operation, our DLU can be easily and seamlessly integrated into existing deep neural networks with different purposes, leading to better performance, but with few additional parameters.

\item \textbf{Scalability}. As an upsampling operation with strong scalability, the proposed DLU is quite suitable for deep networks with aggressive upsampling ratios which are becoming increasingly common. For example, Upernet \cite{xiao2018unified}, Libra RCNN \cite{pang2019libra} utilizes 16$\times$, 8$\times$ upsampling when combining feature maps with different resolutions in object detection;  FCN \cite{long2015fully} uses 32$\times$, 16$\times$, 8$\times$ upsampling in its FCN-8s version in semantic segmentation; RCAN \cite{zhang2018image}, SAN \cite{dai2019second} uses 8$\times$ upsampling in their upscale module for super-resolution.

\end{itemize}

We hope that our DLU can be widely adopted and utilized in different tasks, not only improving the overall performance but also keeping the model complexity stable, and making a positive contribution to the field.

\section{CONCLUSIONS}

{In this paper, we propose a lightweight upsampling operation termed Dynamic Lightweight Upsampling (DLU). The key idea is to avoid generating a large collection of independent kernels with massive trainable parameters, but sample in a source kernel space to reduce redundancy. Extensive experimental results on three mainstream vision tasks with four representation backbones show that the proposed DLU can be seamlessly integrated into existing networks, and effectively improve the performance of downstream tasks.}

\subsubsection{Acknowledgements} This work was supported by the National Natural Science Foundation of China under Grant 62001482 and Hunan Provincial Natural Science Foundation of China under Grant 2021JJ40676.

%
%
%
\bibliographystyle{splncs04}
\bibliography{mybibfile}

\clearpage
\begin{appendix}

\section{Proof}\label{secA1}
Here, we provide proof that all the kernels in the expanded kernel space of our DLU would be softmax normalized after the normalization of the source kernel space. To prove this, we first prove that an interpolated kernel at an arbitrary position $(i',j')$ in the source kernel space is softmax normalized.

Given a new kernel located at position $(i',j')$ in the source kernel space, and surrounded by four source kernels at position $(x_{l},y_{t})$, $(x_{r},y_{t})$, $(x_{l},y_{b})$, $(x_{r},y_{b})$, where $x_{l}=\lfloor i' \rfloor$, $x_{r}=\lceil i' \rceil$, $y_{t}=\lfloor j' \rfloor$, $y_{b}=\lceil j' \rceil$. Since the kernel is sampled from the source kernel space using bilinear interpolation channel-wisely, its response can be written as:
\begin{align}
\label{eq:eq2}
&\forall c: f_c(i',j') \\
&= w_y(w_xf_c(x_{l},y_{t})+(1-w_x)f_c(x_{r},y_{t}))  \nonumber \\
    &+  (1-w_y)(w_xf_c(x_{l},y_{b})+(1-w_x)f_c(x_{r},y_{b}))  \nonumber
\end{align}%
\noindent where $f_c(i',j')$ represents response of the $c$-th channel of the kernel at position $(i',j')$, $w_x=\lceil i' \rceil-i'$, $w_y=\lceil j' \rceil-j'$.

Since the source kernel space is normalized, hence we can get: 
\begin{align}
    \forall c: f_c(x_{l},y_{t})>0,\sum_{c=0}^{k_{up}^2-1}f_c(x_{l},y_{t}) =1 \nonumber  \\
    \forall c: f_c(x_{r},y_{t})>0,\sum_{c=0}^{k_{up}^2-1}f_c(x_{r},y_{t}) =1 \nonumber  \\
    \forall c: f_c(x_{l},y_{b})>0,\sum_{c=0}^{k_{up}^2-1}f_c(x_{l},y_{b}) =1 \nonumber  \\ 
    \forall c: f_c(x_{r},y_{b})>0,\sum_{c=0}^{k_{up}^2-1}f_c(x_{r},y_{b}) =1 
\label{eq:eq3}
\end{align}%
\noindent where $k_{up}^2$ is the total channel number.

Considering $0\leq w_x\leq 1$, $0\leq w_y\leq 1$, combine Equation \ref{eq:eq2} and Equation \ref{eq:eq3}, it can be easily proved:
\begin{equation}
    \forall c: f_c(i',j')\ge0
\label{eq:eq4}
\end{equation}%

Further, we can have:

\begin{align}
    & \sum_{c=0}^{k_{up}^2-1}f_c(i',j') \nonumber \\
     =& w_y(w_x\sum_{c=0}^{k_{up}^2-1}f_c(x_{l},y_{t})+(1-w_x)\sum_{c=0}^{k_{up}^2-1}f_c(x_{r},y_{t}))  \nonumber \\
    + & (1-w_y)(w_x\sum_{c=0}^{k_{up}^2-1}f_c(x_{l},y_{b})+(1-w_x)\sum_{c=0}^{k_{up}^2-1}f_c(x_{r},y_{b}))  \nonumber \\
     =&w_y(w_x+(1-w_x))+(1-w_y)(w_x+(1-w_x))  \nonumber \\
     =&1
\label{eq:eq5}
\end{align}%

{Combining the results of Equation \ref{eq:eq4} and Equation \ref{eq:eq5}, the new kernel at arbitrary positions in the source kernel space has been proved to be softmax normalized.}

For our DLU, its kernels are sampled from the source kernel space. Considering it has already been proved that arbitrary interpolated kernel in the source kernel space is softmax normalized, the kernels in the expanded kernel space of our DLU are also softmax normalized.

\section{Performance on Instance Segmentation}

For instance segmentation, we take the Mask RCNN \cite{He2017Mask} as the baseline framework and then compare the segmentation performance of different upsampling operations on the MS COCO dataset. Results are measured with mask mAP. The calculation of mask mAP is the same as the mAP used in object detection, except that mask AP is evaluated on the IoU between predictions and ground truth masks, rather than the IoU between two boxes.

Mask R-CNN adopted a deconvolution layer to upsample the RoI features in its mask prediction branch. In the following experiments, we simply replaced the deconvolution layer with CARAFE and DLU, and evaluate the segmentation performance in the case of 2$\times$ and 4$\times$ upsampling ratio.

{The comparative results achieved on the MS COCO dataset are reported in Table \ref{table5}. It can be seen that CARAFE and DLU achieve identical performance when the upsampling ratio is set to 2, but DLU achieves better performance as the upsampling rate increase to 4. Moreover, DLU always introduces fewer extra parameters (47\% for 2$\times$ and only 20\% for 4$\times$), further illustrating the superiority of our method on different tasks. In addition to quantitative performance, we also present several instance segmentation results achieved by our proposed DLU model and other baseline methods in Fig. \ref{fig:fig11}.

\begin{table}[t]
\caption{Instance segmentation results on MS COCO 2017 val with Mask RCNN \cite{He2017Mask}. Various upsampling methods are used in different upsample ratio.}\label{table5}
\centering
\tabcolsep=0.3cm
\footnotesize\begin{tabular}{cccc}
\Xhline{2.0\arrayrulewidth}
{Mask head} & {Method} & {Mask mAP} & {\textbf{Params}} \\
\Xhline{2.0\arrayrulewidth}
\multirow{5}{*}{$\times$2} 
    & {Nearest} & {33.3} & {0} \\
    & {Deconv} & {34.6} & {262K} \\
    &{CARAFE} & {\bf 35.0} & {74K} \\
    &{DLU} & {\bf 35.0} & {35K} \\
\hline
\multirow{5}{*}{$\times$4} 
    & {Nearest} & {32.4} & {0} \\
    & {Deconv} & {34.6} & {1.0M} \\
    &{CARAFE} & {34.8} & {247K} \\
    &{DLU}  & {\bf 35.0} & {49K} \\
\Xhline{2.0\arrayrulewidth}
\end{tabular}
\end{table}

\begin{figure}[t]
\centering
\centerline{\includegraphics[width=12cm]{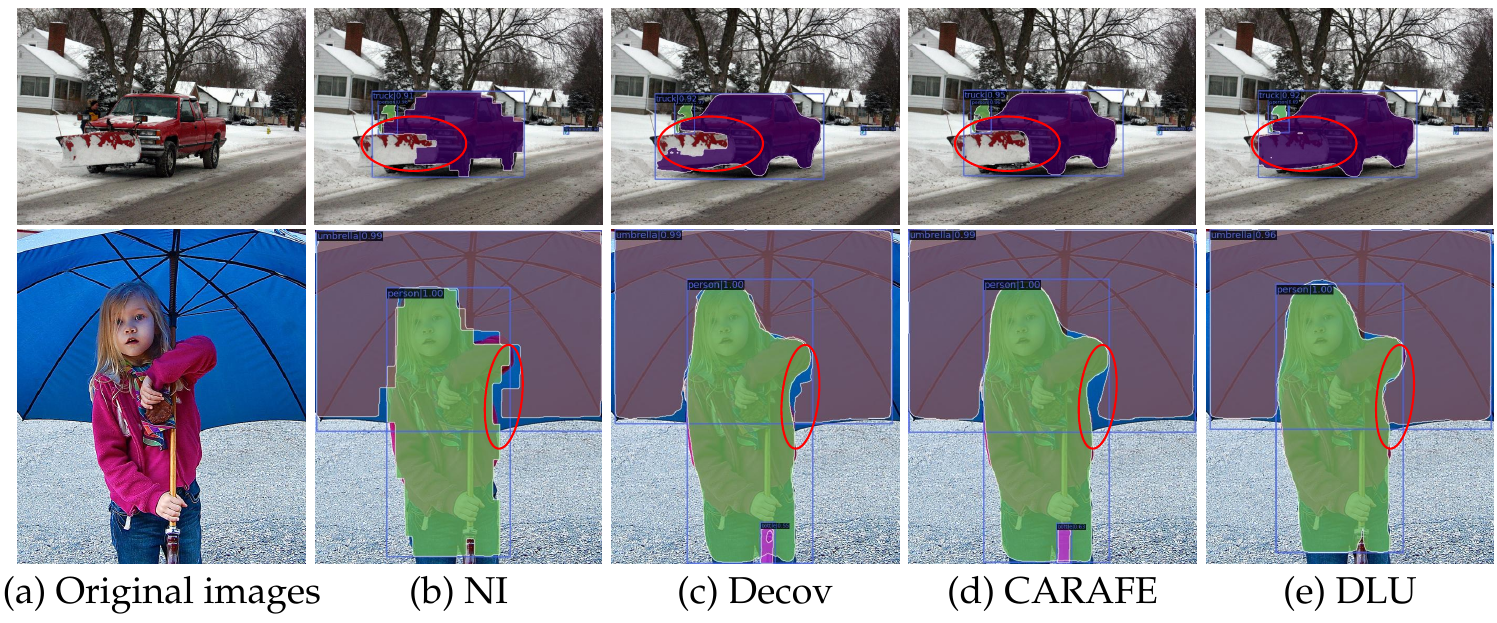}}
\caption{Object detection and instance segmentation results of Mask RCNN with various upsampling operations in its mask head.}
\label{fig:fig11}
\end{figure}

\section{Ablation Study}
{In this section, we conduct ablation studies to analyze the impact of hyperparameters in the proposed DLU operation. Specifically, we take FPN with ResNet-50 backbone in the following ablation experiments on the MS COCO dataset \cite{lin2014microsoft}.}

\begin{table}[t]

\caption{Detection results with various encoder kernel size $k_{encoder}$ and receptive field for upsampling $k_{up}$.}\label{table6}
\centering
\tabcolsep=0.3cm
\footnotesize\begin{tabular}{ccccc}
\Xhline{2.0\arrayrulewidth}
$k_{encoder}$ &$k_{up}$ & AP & AP$_{50}$ & AP$_{75}$\\
\Xhline{2.0\arrayrulewidth}
{1} & {3} &{38.3} & {59.3} & {41.6}\\
{1} & {5} &{38.5} & {59.7} & {41.4}\\
{3} & {3} &{37.9} & {59.0} & {40.9}\\
{3} & {5} &{\bf 38.6} & {\bf 59.9} & {\bf 41.8}\\
{3} & {7} &{38.5} & {59.7} & {41.7}\\
{5} & {3} &{38.1} & {59.1} & {41.1}\\
{5} & {5} &{38.3} & {59.6} & {41.4}\\
{5} & {7} &{\bf 38.6} & {\bf 59.9} & {\bf 41.8}\\
\Xhline{2.0\arrayrulewidth}
\end{tabular}
\end{table}

\begin{table}[t]
\caption{Ablation study of various compressed channels $C_m$.}\label{table7}
\centering
\tabcolsep=0.3cm
\footnotesize\begin{tabular}{rcccccc}
\Xhline{2.0\arrayrulewidth}
$C_m$ & AP & AP$_{50}$ & AP$_{75}$ & AP$_{S}$ & AP$_{M}$ & AP$_{L}$\\
\Xhline{2.0\arrayrulewidth}
{32} & {38.4} & {59.6} & {41.3} & {21.8} & {41.9} & {49.8} \\
{64} &{\bf 38.6} & {\bf 59.9} & {\bf 41.8} & {\bf 22.6} & {\bf 42.2} & {\bf 50.1} \\
{128} & {38.3}& {59.6}& {41.2}& {22.2}& {42.0}& {49.6} \\
{256} & {38.2}& {59.5}& {41.4}& {22.1}& {41.8}& {49.2}  \\
\Xhline{2.0\arrayrulewidth}
\end{tabular}
\end{table}

\smallskip\noindent\textbf{Varying kernel size of the encoder and upsampling.} {We investigate the impact of $k_{encoder}$ and $k_{up}$ to the detection performance. As shown in Table \ref{table6}, it seems better performances are more likely to be achieved when $k_{encoder}$ is smaller than $k_{up}$. Additionally, when $k_{encoder}$ is fixed, larger $k_{up}$ seems to have better performance. Considering both the performance and efficiency, we finally set $k_{up}=5$ and $k_{encoder}=3$ in all experiments.}

\smallskip\noindent\textbf{Varying channel number in the channel compressor.} We further explore the influence of $C_m$ in the channel compressor. In fact, there has been no exact solution for how determining the number of nodes in the channel compressor. However, there are several empirically derived schemes \cite{heaton2008introduction}. Among them, the most commonly relied on is ``the optimal size of the hidden layer is usually between the size of the input and size of the output layers''.

For our case in the FPN \cite{lin2017feature}, the channel number of the input feature map is 256, while the channel number of the output feature map is 25 ($k_{up}^2$). Therefore, we conduct several experiments with different $C_m$ settings between 25 and 256. Performance is shown in Table \ref{table7}.

By changing the number of channels $C_m$ from 32 to 256, we find the best performance is achieved when $C_m$ = 64 in the channel compressor (Table \ref{table7}). Therefore, we choose $C_m$ = 64 by default in our experiments.
}

\end{appendix}

\end{document}